\documentclass[a4,10pt]{article}

\usepackage[margin=3cm]{geometry}
\usepackage[numbers]{natbib}
\usepackage{booktabs}
\usepackage[margin=5mm,font=small]{caption}
\usepackage{subcaption}
\usepackage{graphicx}
\usepackage{slashbox}
\usepackage{amsmath,amssymb}

\usepackage{calc}
\usepackage{color}
\newcommand{\gray}[1] {\textcolor[gray]{0.25}{\textbf{#1}}}

\graphicspath{{pics/}}

\title{Efficient Defenses Against Adversarial Attacks}
\author{Valentina Zantedeschi, Maria-Irina Nicolae, Ambrish Rawat\\IBM Research Ireland}
\date{}

\begin{document}
\maketitle

\begin{abstract}
  Following the recent adoption of deep neural networks (DNN) accross a wide range of applications, adversarial attacks against these models have proven to be an indisputable threat.
  Adversarial samples are crafted with a deliberate intention of undermining a system.
  In the case of DNNs, the lack of better understanding of their working has prevented the development of efficient defenses.
  In this paper, we propose a new defense method based on practical observations which is easy to integrate into models and performs better than state-of-the-art defenses.
  Our proposed solution is meant to reinforce the structure of a DNN, making its prediction more stable and less likely to be fooled by adversarial samples.
  We conduct an extensive experimental study proving the efficiency of our method against multiple attacks, comparing it to numerous defenses, both in white-box and black-box setups.
  Additionally, the implementation of our method brings almost no overhead to the training procedure, while maintaining the prediction performance of the original model on clean samples.
\end{abstract}

\section{Introduction}

Deep learning has proven its prowess across a wide range of computer vision applications, from visual recognition to image generation~\citep{lecun2015deep}.
Their rapid deployment in critical systems, like medical imaging, surveillance systems or security-sensitive applications, mandates that reliability and security are established \emph{a priori} for deep learning models.
% This is even more imperative in the light of recent advances~\citep{zhang2016generalize} which have exposed some critical vulnerabilities of deep learning.
Similarly to any computer-based system, deep learning models can potentially be attacked using all the standard methods (such as denial of service or spoofing attacks), and their protection only depends on the security measures deployed around the system.
Additionally, DNNs have been shown to be sensitive to a threat specific to prediction models: \emph{adversarial examples}.
These are input samples which have deliberately been modified to produce a desired response by a model (often, misclassification or a specific incorrect prediction which would benefit the attacker).

% More precisely, a deployed model can be fooled when making predictions for \emph{adversarial examples}.
% One in particular, namely \emph{adversarial examples}, opens the possibility of targeted attacks by adversaries seeking benefit.

Adversarial examples pose an asymmetric challenge with respect to attackers and defenders.
An attacker aims to obtain his reward from a successful attack without raising suspicion.
% incentivised to cause maximum damage while maintaining inconspicuousness when crafting adversarial inputs.
% Furthermore, an attacker is also motivated to achieve the desired output with minimal effort.
A defender on the other hand is driven towards developing strategies which can guard their models against all known attacks and ideally for all possible inputs.
Furthermore, if one would try to prove that a model is indeed secure (that is, can withstand attacks yet to be designed), one would have to provide formal proof, say, through verification~\citep{huang2016}.
This is a hard problem, as this type of methods does not scale to the number of parameters of a DNN.
For all these reasons, finding defense strategies is hard.

\begin{figure}[h]
  \centering
  ~~~~~~~~~~~~\includegraphics[width=.45\textwidth]{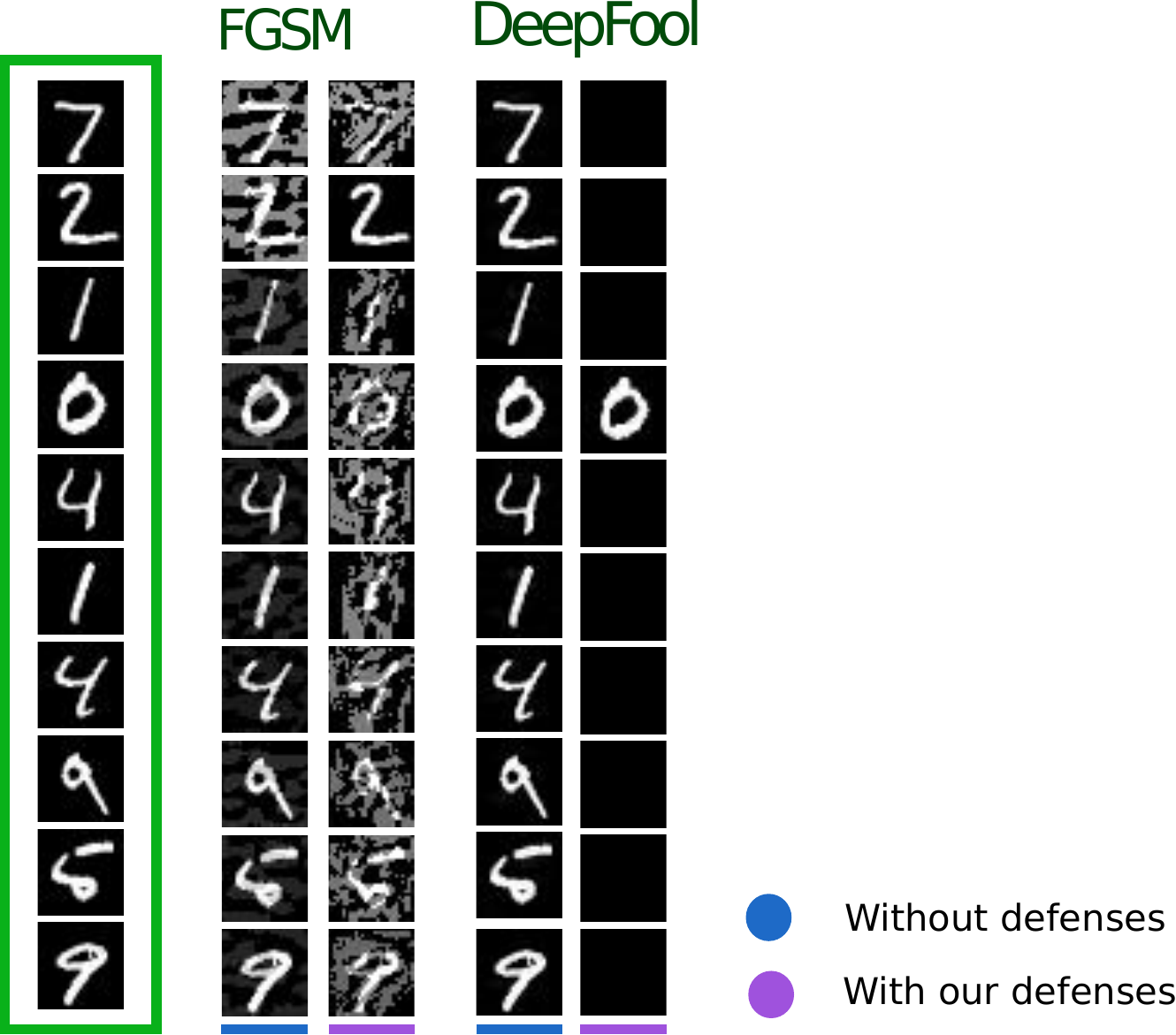}
  \caption{Minimal perturbations needed for fooling a model on the first ten images from MNIST. The original examples are marked by the green rectangle. With our defenses, the attack becomes visually detectable.}
  \label{fig:min_perts}
\end{figure}

% This new kind of attacks differs from the historical setting of learning with malicious error~/citep{kearns1993learning} because they can damage the system even after a prediction model have been learned.

Apart from the security aspects, adversarial examples for image classification have other peculiar properties.
First and foremost, the imperceptible difference between adversarial and legitimate examples provides them with an effortless capacity of attack; this is only reinforced by the transferability of such samples across different models, allowing for black-box attacks.
Moreover, the high confidence of misclassification proves that the model views adversarial samples as regular inputs.
The potential damage of adversarial attacks is increased by the existence of such samples in the physical world~\citep{DBLP:journals/corr/KurakinGB16}: showing printed versions of adversarial images to a camera which will feed then to a DNN has the same effect as just presenting them to the model without passing through a physical support.
Another intriguing property is that nonsensical inputs (e.g. crafted noise) are interpreted by the model as natural examples with high confidence~\citep{nguyen2014}.
From the perspective of learning, these counterintuitive aspects are related to the more fundamental questions of what does a model learn and how well it is able to generalize~\citep{zhang2016generalize,madry2017}.
As a matter of fact, the mere existence of adversarial samples suggests that DNNs might be failing to extract concepts from the training set and that they would, instead, memorize it.

% Other types of adversarial: image segmentation~\citep{metzen2017image_seg}, speech recognition~\citep{197215}, detection systems~\citep{}, etc.
A good understanding of the weaknesses of deep learning models should bring to better attack strategies, and most importantly more effective defenses.
While the cause is not completely understood to this day, multiple hypotheses have been suggested to explain their sensitivity to adversarial inputs.
One of the first such hypothesis stated that the high complexity and non-linearity~\citep{szegedy2013} of neural networks can assign random labels in areas of the space which are under-explored.
Moreover, these adversarial examples would be rare and would only exist in small regions of the space, similar to pockets.
These theories have been refuted, since they are unable justify the transferability of adversarial samples from one model to another.
Moreover, it has been shown that linear models also suffer from this phenomenon.
\citet{goodfellow2014fgsm} proposed a \emph{linearity hypothesis} instead: deep neural networks are highly non-linear with respect to their parameters, but mostly linear with respect to their inputs, and adversarial examples are easy to encounter when exploring a direction orthogonal to a decision boundary.
Another conjecture to explain the existence of adversarial examples is the cumulation of errors while propagating the perturbations from layer to layer.
A small carefully crafted perturbation in the input layer may result in a much greater difference in the output layer, effect that is only magnified in high dimensional spaces, causing the activation of the wrong units in the upper layers.

In this paper, we make the following contributions:

\begin{itemize}
  \item We propose a two-fold defense method which is easy to setup and comes at almost no additional cost with respect to a standard training procedure.
  The method is designed to reinforce common weak points of deep networks and to smooth the decision functions.
  As a consequence, it is agnostic to the type of attack used to craft adversarial examples, making it effective in multiple settings.
  Figure~\ref{fig:min_perts} shows examples of clean images which are then perturbed with standard attacks.
  When the model uses the proposed defense, the perturbation necessary for misclassification is much larger, making the attack detectable and, in some cases, turning the images into nonsense.

  \item We perform an extended experimental study, opposing an important number of attacks to the most effective defense methods available, alongside our proposed defense.
  We evaluate them according to multiple metrics and prove that accuracy by itself is not a sufficient score.
  We explore white-box and black-box attacks alike, to account for different adversarial capacities, as well as transferability of examples.
\end{itemize}

The rest of this document is structured as follows.
Section~\ref{sec:related} provides an overview of the existing attack and defense methods.
Section~\ref{sec:main} introduces the proposed defense method, followed by an extensive experimental study in Section~\ref{sec:experiments}.
We conclude this paper in Section~\ref{sec:conclusion}.

\section{Related Work}
\label{sec:related}

\paragraph{Background and notation}

A neural network can be formalized as a function $F(x, \theta) = y$ taking inputs $x \in \mathbb{R}^n$ and providing outputs $y \in \mathbb{R}^m$ w.r.t. on a set of parameters $\theta$.
This paper covers the case of multi-class classification, where the last layer in the network is the \emph{softmax} function, and $m$ is the number of labels.
The \emph{softmax} function has two main properties: (i) it amplifies high values, while diminishing smaller ones, and (ii) it outputs vectors $y$ of non-negative values which sum to 1, giving them an interpretation of probability distributions.
The input $x$ is then attributed the class label with the highest probability.
Now consider a network $F$ containing $L$ layers, $F_1$ being the input and $F_L$ the \emph{softmax} layer.
An internal layer $F_j$ can be written as:
\[
  F_j(x_{j-1}, (\theta_j, \hat{\theta_j})) = \sigma(\theta_j \cdot x_{j-1}) + \hat{\theta_j},
\]
where $\sigma(.)$ is the activation function, $\theta_j$ and $\hat{\theta_j}$ are the parameters of layer $F_j$ and $x_{j-1}$ is the output of the previous layer.
The most popular activation function, and almost the only one that has been studied in an adversarial setting, is the Rectified Linear Unit, or RELU~\citep{glorot2011deep}.
% TODO activation functions

In our study of adversarial learning, we focus on the task of image classification, as this type of data is readily interpretable by humans: it is possible to distinguish true adversarial examples, i.e.\ perceptually identical to the original points, from rubbish ones, i.e.\ overly perturbed and meaningless.
% providing an adversarial dimension to attacks as opposed to just random noise.
Consider an image of size $w \times h$ pixels with pixel values scaled between 0 and 1.
In the greyscale case, such an image can be viewed as a vector $x \in \mathbb{R}^{w \cdot h}$, where each $x_i$ denotes a pixel.
Similarly, RGB images have three color channels and are written as vectors $x \in \mathbb{R}^{3 \cdot w \cdot h}$.
The problem of crafting adversarial examples can be formulated as trying to find samples $x' = x + \Delta x$ which fool the model into making incorrect predictions.
\citet{szegedy2013} is one of the first adversarial attacks against deep neural networks, aiming to achieve a target class prediction.
Their method is expressed as a box-constrained optimization objective, solved through L-BFGS:

\begin{equation}
  \min_{\Delta x} F(x + \Delta x) \neq F(x), \quad
  \mbox{s.t.} \quad x' \in [0, 1]^n.
  \label{eq:lbfgs}
\end{equation}

Although effective in producing adversarial examples, this attack is computationally expensive to the point where its usage is not practical.
In the aim of speeding up the computation of adversarial examples, the attacking methods subsequently proposed~\citep{papernot2015,goodfellow2014fgsm,DBLP:journals/corr/Moosavi-Dezfooli15} all solve a first-order approximation of this original problem, which has a geometrical interpretation.
The resulting perturbations are also effective in fooling the model, probably because commonly used deep architectures, such as the ones with piecewise linear RELU, are highly linear w.r.t. the input ~\citep{goodfellow2014fgsm}.
% give a mathematical formulation of the new problem

We now discuss the methods which have shaped the current state-of-the-art in adversarial attacks for deep neural networks.
Any of the following attacks can be deployed in two ways: either by crafting the adversarial examples having knowledge of the architecture and the parameters of the attacked model (white-box attacks) or by crafting them using a similar model, plausible for the considered task, or a surrogate one as in~\citet{Papernot2017} without exploiting any sensitive information (black-box attacks).

\paragraph{Attacks}
One of the gradient-based methods is the Jacobian saliency map attack (JSMA)~\citep{papernot2015}, which uses the derivative of the neural network with respect to the input image to compute the distortion iteratively.
Each iteration, the pixel with the highest derivative is modified by a fixed value (the budget for the attack), followed by recomputing the saliency map, until the prediction has changed to a target class.
The adversarial images produced by JSMA are subtle and effective for attacks, but they still require an excessive amount of time to compute.

The fast gradient sign method (FGSM)~\citep{goodfellow2014fgsm} has been introduced as a computationally inexpensive, but effective alternative to JSMA.
FGSM explores the gradient direction of the cost function and introduces a fixed amount of perturbation to maximize that cost.
In practice, the examples produced by this attack are more easily detectable and require a bigger distortion to achieve misclassification than those obtained from JSMA.
An iterative version of FGSM, where a smaller perturbation is applied multiple times, was introduced by~\citet{DBLP:journals/corr/KurakinGB16}.

Instead of using a fixed attack budget as for the last two methods, DeepFool~\citep{DBLP:journals/corr/Moosavi-Dezfooli15} was the first method to compute and apply the minimal perturbation necessary for misclassification under the $L_2$ norm.
The method performs iterative steps on the adversarial direction of the gradient provided by a locally linear approximation of the classifier.
Doing so, the approximation is more accurate than FGSM and faster than JSMA, as all the pixels are simultaneously modified at each step of the method, but its iterative nature makes DeepFool computationally expensive.
In~\citep{Moosavi-Dezfooli16adversarial,Moosavi-Dezfooli2017}, the authors extend DeepFool in order to craft a \emph{universal perturbation} to be applied indifferently to any instance: a fixed distortion is computed from a set of inputs, allowing to maximize the predictive error of the model on that sample.
The perturbation is computed by a greedy approach and needs multiple iterations over the given sample before converging.
To the extent where the sample is representative to the data distribution, the computed perturbation has good chances of achieving misclassification on unseen samples as well.
% The universal perturbation is a generic strategy which needs a concrete attack method to compute the perturbation; the authors choose DeepFool for this purpose.

One method aiming to compute good approximations of Problem~\eqref{eq:lbfgs} while keeping the computational cost of perturbing examples low has been proposed in~\citet{carlini2017}.
The authors cast the formulation of \citet{szegedy2013} into a more efficient optimization problem, which allows them to craft effective adversarial samples with low distortion.
They define three similar targeted attacks, based on different distortion measures: $L_2$, $L_0$ and $L_{\infty}$ respectively.
In practice, even these attacks are computationally expensive.
% what's the difference with the other methods? should we cite it?

If it is difficult to find new methods that are both effective in jeopardizing a model and computationally affordable, defending from adversarial attacks is even a harder task.
On one hand, a good defense should harden a model to any known attack and, on the other hand, it should not compromise the discriminatory power of the model.
In the following paragraph, we report the most effective defenses proposed for tackling adversarial examples.

\paragraph{Defenses}
A common technique for defending a model from adversarial examples consists in augmenting the training data with perturbed examples (technique known as `adversarial training`~\citep{szegedy2013}) by either feeding a model with both true and adversarial examples or learning it using the modified objective function:
$$\hat{J}(\theta,x,y) = \alpha J(\theta,x,y) + (1 - \alpha)J(\theta,x+\Delta x,y)$$
with $J$ the original loss function.
The aim of such defense is to increase the model's robustness in specific directions (of the adversarial perturbation) by ensuring that it will predict the same class for the true example and its perturbations along those directions.
In practice, the additional instances are crafted for the considered model using one or multiple attack strategies, such as FGSM~\citep{goodfellow2014fgsm}, DeepFool~\citep{DBLP:journals/corr/Moosavi-Dezfooli15} and virtual adversarial examples~\citep{miyato2017virtual}.

However, adversarially training a model is effective only on adversarial examples crafted on the original model, which is an improbable situation considering that an attacker might not have access to exactly the same model for computing the perturbations.
% and is useless against new attacks, on which the model has not been trained.
Additionally, adversarial training has been proved to be easily bypassed through a two-step attack~\citep{tramer2017ensemble}, which first applies a random perturbation to an instance and then performs any classical attack technique.
The success of this new attack, and of black-box attacks in general, is due to the sharpness of the loss around the training examples: if smoothing the loss in few adversarial directions makes  ineffective gradient-based attacks on those directions, it also makes the loss sharper in the other directions, leaving the model more vulnerable through new attacks.

Unlike adversarial training, a different family of defenses aims to increase the robustness of deep neural networks to adversarial examples independently of the attack.
Among these \emph{attack-agnostic} techniques, we find defensive distillation~\citep{papernot2015distill,papernot2016distill,carlini2016distillation}, which hardens the model in two steps: first, a classification model is trained and its \emph{softmax} layer is smoothed by division with a constant $T$; then, a second model is trained using the same inputs, but instead of feeding it the original labels, the probability vectors from the last layer of the first model are used as soft targets.
The second model is then used for future deployment.
The advantage of training the second model with this strategy is that it makes for a smoother loss function.
It has been shown in~\citet{hazan2016perturbation} that a similar behavior can be obtained at a cheaper cost by training a model using smoothed labels.
This technique, called \emph{label smoothing}, involves converting class labels into soft targets (value close to 1 for the target class and the rest of the weight distributed on the other classes) and use these new values for training the model instead of the true labels.
As a consequence, one saves the needs to train an additional model as for defensive distillation.
% The effectiveness of the attacks which were available at the time this defense was designed was almost reduced to zero.
% However, \citet{carlini2016distillation} have shown that distillation does not protect the model against their attacks~\citep{carlini2017}.

Another model hardening technique is feature squeezing~\citep{xu2017feature_squeeze,xu2017feature_squeeze_carlini}.
Its reduces the complexity of the representation of the data so that the adversarial perturbations disappear due to low sensitivity.
The authors propose two heuristics for dealing with images: reducing color depth on a pixel level, that is encoding the colors with less values, and using a smoothing filter over the image.
As an effect, multiple inputs are mapped to the same value, making the model robust to noise and adversarial attacks.
Although this has the collateral effect of worsening the accuracy of the model on true examples, to the best of our knowledge, feature squeezing is the most effective defense to adversarial attacks to date.
% This method has shown efficiency~\citep{xu2017feature_squeeze_carlini} against the attacks from \citet{carlini2017}, but has since been proven to have decreased performance in the case of an adaptive attacker~\citep{carlini2017adaptive}.

% Virtual adversarial training (VAT)~\citep{miyato2017virtual} proposes to replace the hard target class used in adversarial training when generating perturbations with an estimate of the conditional probability of the output labels given the input.
% The computed distortion maximizes a distance measure between the distribution of the outputs for the original and adversarial input.
% VAT has the property of being suitable for a semi-supervised setting, as it does not require input labels for computing the perturbation used in training.

A different approach for protecting models against adversarial attacks are detection systems.
To this end, a certain number of directions have been explored, such as performing statistical tests~\citep{grosse2017}, using an additional model for detection~\citep{gong2017,metzen2017detection} or applying dropout~\citep{srivastava2014} at test time~\citep{feinman2017}.
However, with adversarial examples being relatively close to the original distribution of the data, it has been shown that many detection methods can be bypassed by attackers~\citep{carlini2017bypass}.

As we have described, defending against adversarial examples is not an easy task, and the existing defense methods are only able to increase model robustness in certain settings and to a limited extent.
With these aspects in mind, we now move on to presenting the proposed defense method.

\section{Efficient Defenses}
\label{sec:main}

In this section, we start by introducing the threat model we consider, before presenting the two aspects of our contribution.

\paragraph{Adversary model}
Attackers can be formalized depending on their degree of knowledge, the ways in which they can tamper with the system, as well as the expected reward.
For the purpose of our contribution, modeling the reward is not required.
In this paper, we consider attackers that only have access to test data and, optionally, the trained model.
They are thus unable to tamper with the training sample, unlike in other contexts, such as learning with malicious error~\citep{kearns1993learning}.
We address different settings, depending on the degree of knowledge of the adversary.
The attacker can gain access to information about the learning algorithm, which can include only the architecture of the system or values of the parameters as well, the feature space and the data which was used for training.
Of course, from the perspective of the attacker, the white-box setup is the most advantageous, making the crafting easier.
% This is why designing a good defense method should either prove that white-box attacks cannot be achieved on the system (which is a hard problem) or be able to defend against them.
A good defense method should be able to sustain the strongest type of attack achievable in practice.
On the other hand, it has been shown that in some cases a black-box attack, when the attacker only has access to the input and output of the model, achieves better results than its white-box counterpart~\citep{Papernot2017}.
% This counterintuitive efficiency of transferred samples can be explained by the gradient masking effect. % TODO check this
We thus consider both black-box and white-box attacks when evaluating our method.

\subsection{Bounded RELU}
\label{sec:brelu}

We now introduce the use of the bounded RELU (BRELU)~\citep{liew2016bounded} activation function for hedging against the forward propagation of adversarial perturbation.
Activation functions present in each node of a deep neural network amplify or dampen a signal depending on the magnitude of the input received.
Traditional image classification models use Rectified Linear Units which are known to learn sparse representations for data~\citep{glorot2011deep}, thereby easing the training process.
Recall that a RELU operation squashes negative inputs to zero, while propagating all positive signals.
Given the arrangement of nodes in a DNN, inputs received by a node in one layer depend on the outputs of the nodes from the previous layers of the architecture.
Naturally, with unbounded units like RELU, a small perturbation in the input signal can accumulate over layers as a signal propagates forward through the network.
For an adversarial perturbation, this can potentially lead to a significant change in the output signal for an incorrect class label.
This would be further amplified by the \emph{softmax} operation in the final layer of a classification network.
To curtail this phenomenon, we propose the use of the bounded RELU activation function, defined as follows:

\begin{equation*}
\label{eq:brelu}
   f_t(x) = \begin{cases}
        0, & x < 0 \\
        x, & 0 \leq x < t\\
        t, & x \geq t.\\
        \end{cases}
 \end{equation*}

The parameter $t$ defines the cut-off point where the function saturates.
In practice, it should be set with respect to the range of the inputs.
Notice that a value too small for $t$ would prevent the forward propagation of the input in the network, reducing the capacity of the model to perform well.
On the other hand, too big a value will perpetuate the same behavior as RELU.

To theoretically prove the interest of this simple modification in the architecture, we compare the additive stabilities of a neural network with RELU activations against BRELU activations, following \citet{szegedy2013}.
For a model using RELU, the output difference between an original point and its perturbation can be upper bounded as follows
$$ \forall x, \Delta x \quad || F(x) - F(x + \Delta x) || \leq M || \Delta x || $$
with $ M = \prod_{j=1}^L M_j$ the product of the Lipschitz constants of each layer.
On the contrary, the output difference in a model with BRELU has a tighter bound for a small enough $t$ and, most importantly, a bound independent from the learned parameters of the layers:
$$ \forall x, \Delta x \quad || F(x) - F(x + \Delta x) || \leq t ||\textbf{1}|| .$$
Employing BRELU activation functions, then, improves the stability of the network.

\subsection{Gaussian Data Augmentation}
\label{sec:gaussian}
The intuition behind data augmentation defenses such as adversarial training is that constraining the model to make the same prediction for a true instance and its slightly perturbed version should increase its generalization capabilities.
Although adversarial training enhances the model's robustness to white-box attacks, it fails to protect effectively from black-box attacks~\citep{tramer2017ensemble}.
This is because the model is strengthened only in few directions (usually, one per input sample), letting it be easily fooled in all the other directions.
Moreover, there is no mechanism for preventing the model from making confident decisions for uncertain regions, i.e.\ parts of the input space not represented by the data samples.
Instead, augmenting the training set with examples perturbed using Gaussian noise, as we propose in this paper, on one hand allows to explore multiple directions and, on the other, smooths the model confidence.
While the former property can be achieved through any kind of noise (e.g.\ uniform noise), the latter is peculiar to using a Gaussian distribution for the perturbations: the model is encouraged to gradually decrease its confidence moving away from an input point.

We thus propose a new formulation of the problem of learning classifiers robust to adversarial examples:

$$ \min_\theta \underset{(x,y) \sim \mathcal{D}}{\mathbb{E}} \:\:\: \underset{\Delta x \sim \mathcal{N}(0,\sigma^2) }{\mathbb{E}} J(\theta, x+\Delta x, y), $$
where $\sigma$ indicatively corresponds to the acceptable non-perceivable perturbation.
The aim is to enforce the posterior distribution $p(y | x)$ to follow $\mathcal{N}(x,\sigma^2)$.
This formulation differs from~\citep{madry2017} in considering all possible local perturbations (and not only the one with the maximal threatening power) and in weighing them with respect to their magnitude.

In the rest of the paper, we will solve a Monte Carlo approximation of the solution of the previous problem, by sampling $N$ perturbations per instance from $\mathcal{N}(0,\sigma^2)$:

$$ \min_\theta \underset{(x,y) \sim \mathcal{D}}{\mathbb{E}} \:\:\: \frac{1}{N} \sum_{i=1}^N J(\theta, x+\Delta x_i, y) $$
that converges almost surely to the true one for $N \to \infty$.
Let's note $\mu = \mathbb{E}_{\Delta x \sim \mathcal{N}(0,\sigma^2)} J(\theta, x+\Delta x, y)$.
After Hoeffding's inequality, the amount of deviation of the empirical approximation from the theoretical one can be quantified for all $t \geq 0$ as

$$ \Pr \left [ \left| \frac{1}{N} \sum_{i=1}^N J(\theta, x+\Delta x_i, y) - \mu \right| \geq t \right ] \leq \exp \left (- \frac{t^2 N}{2 \sigma^2} \right). $$

\begin{figure}
  \begin{subfigure}{\textwidth}
    \centering
      \includegraphics[width=0.32\textwidth]{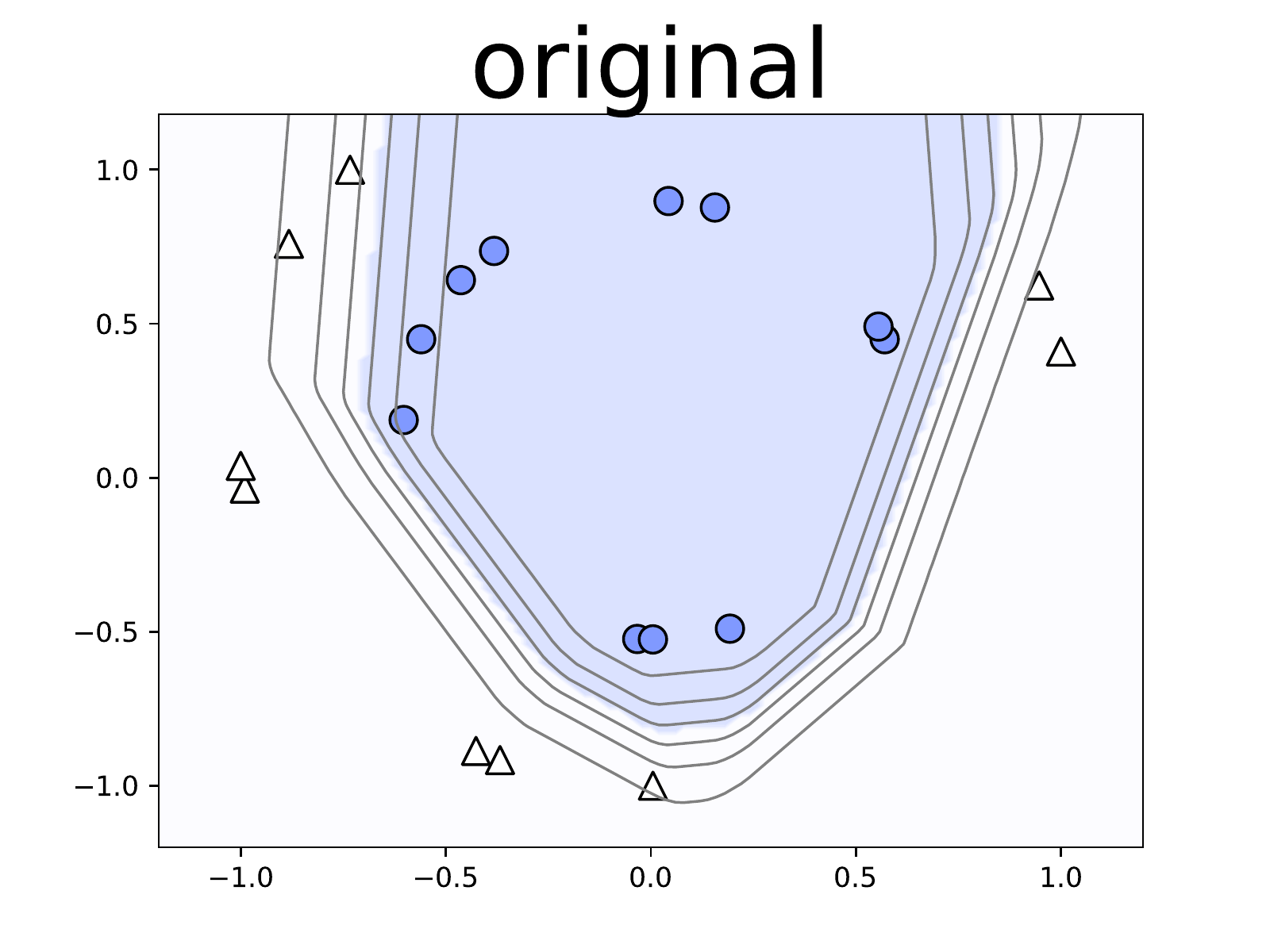}
      \includegraphics[width=0.32\textwidth]{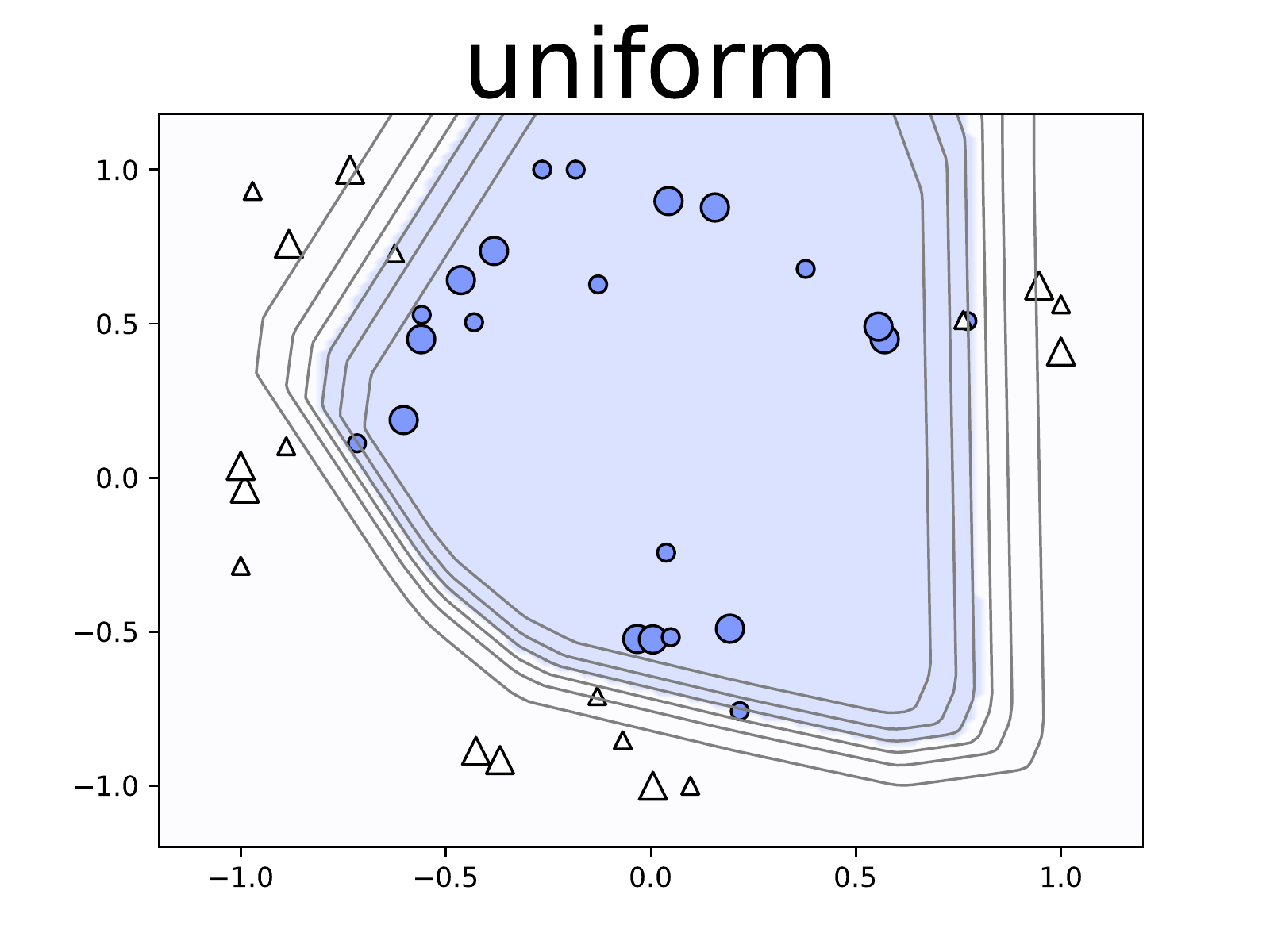}
      \includegraphics[width=0.32\textwidth]{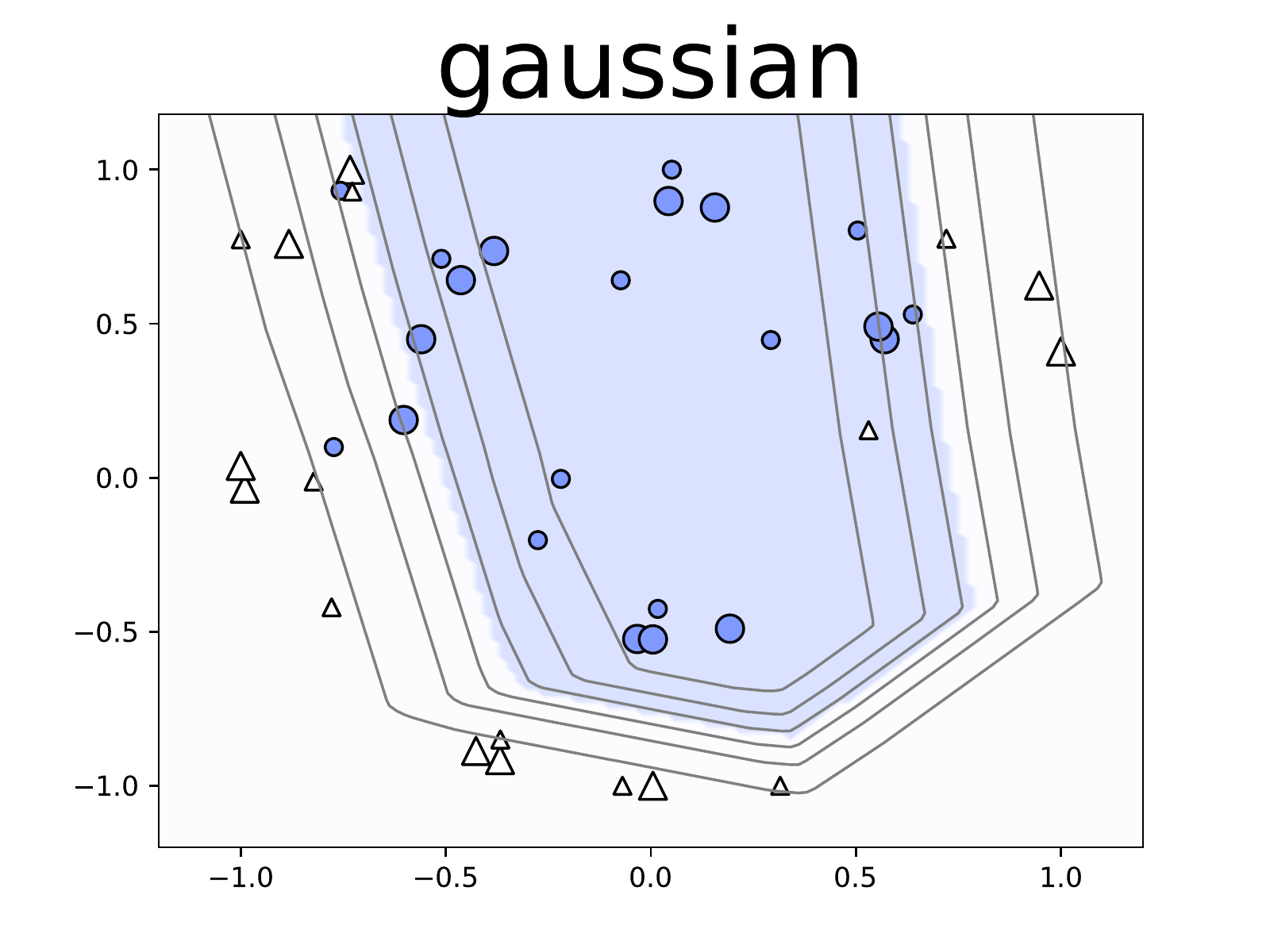}\\
      \includegraphics[width=0.32\textwidth]{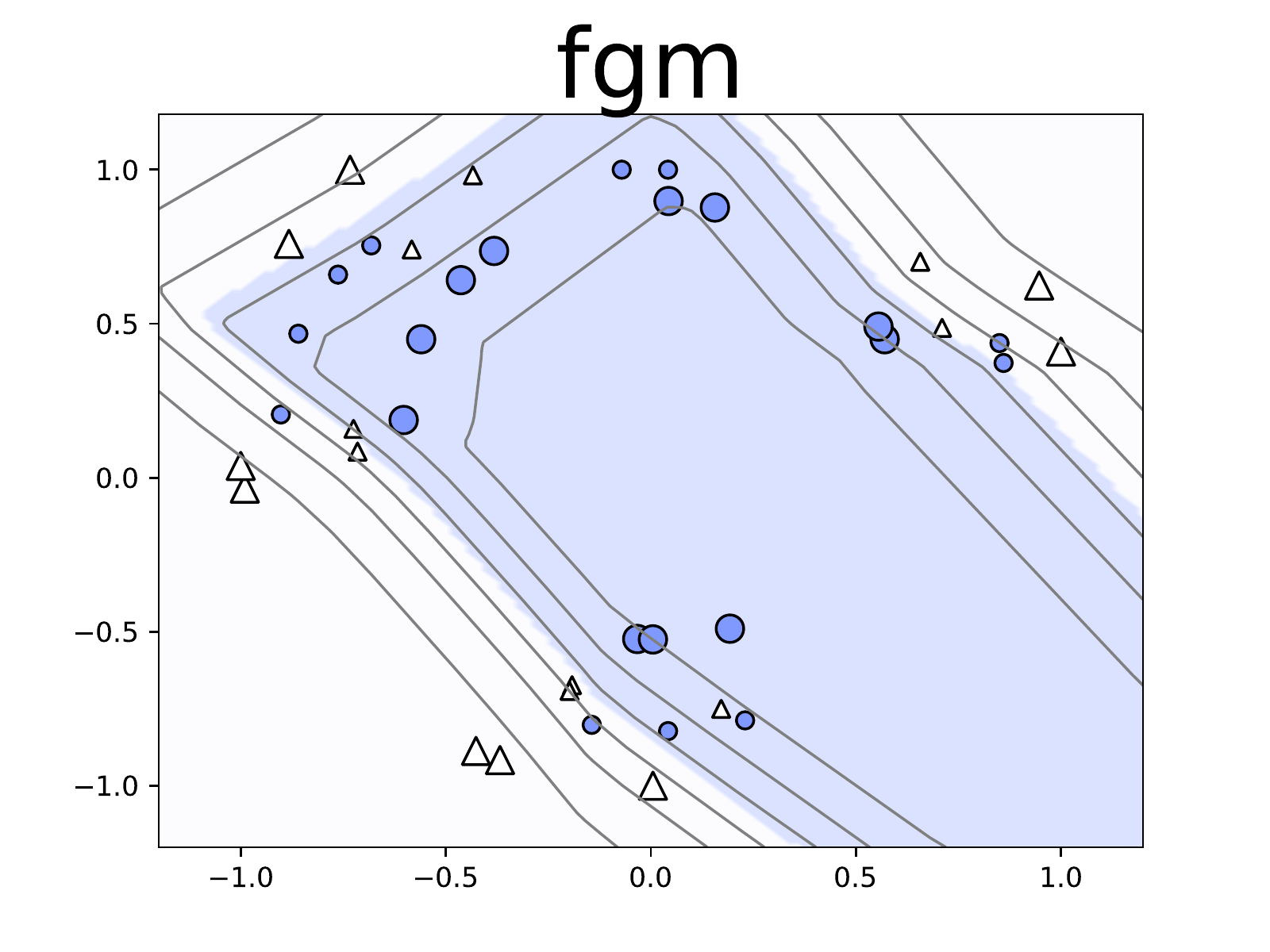}
      \includegraphics[width=0.32\textwidth]{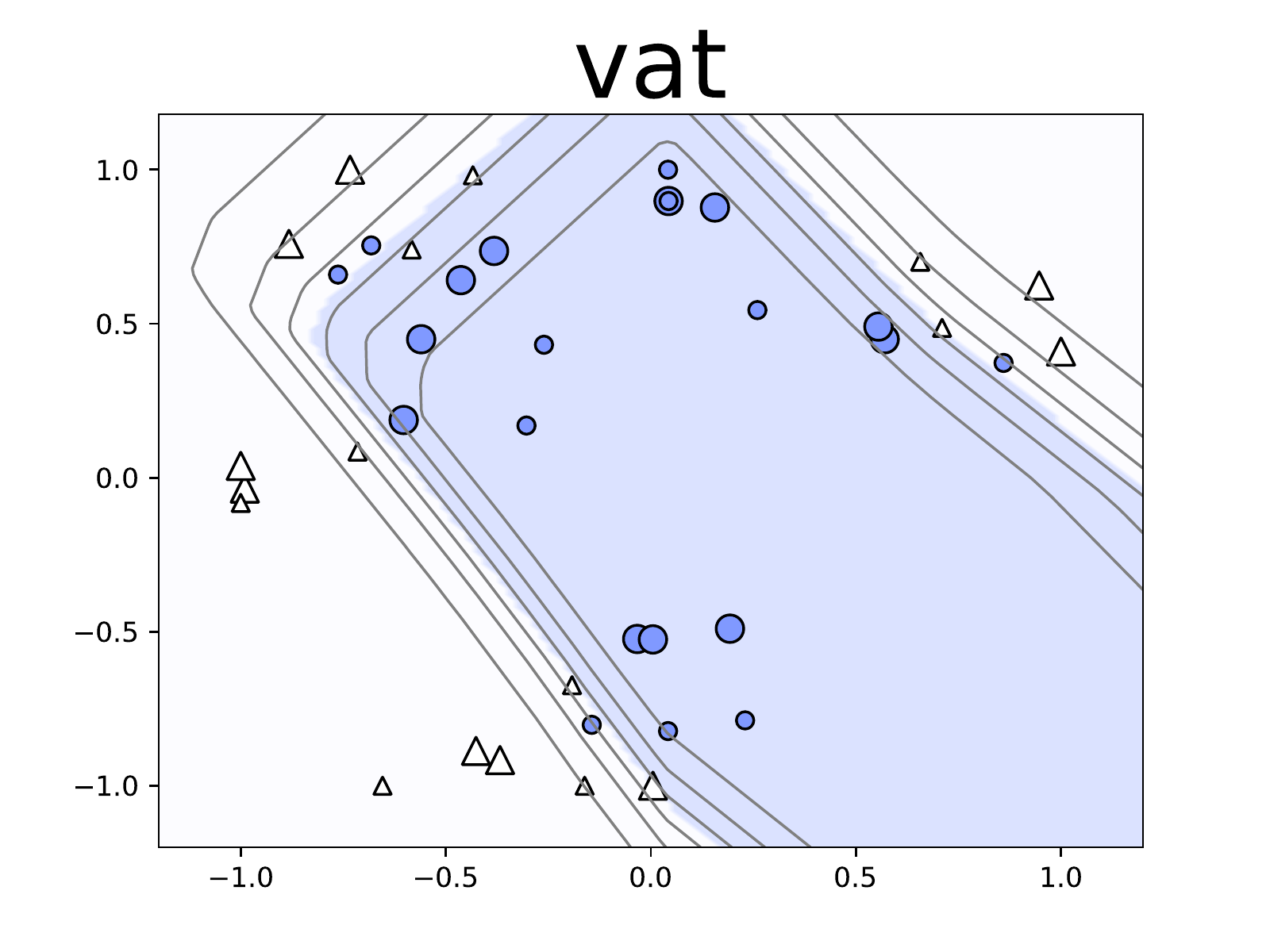}
      \includegraphics[width=0.32\textwidth]{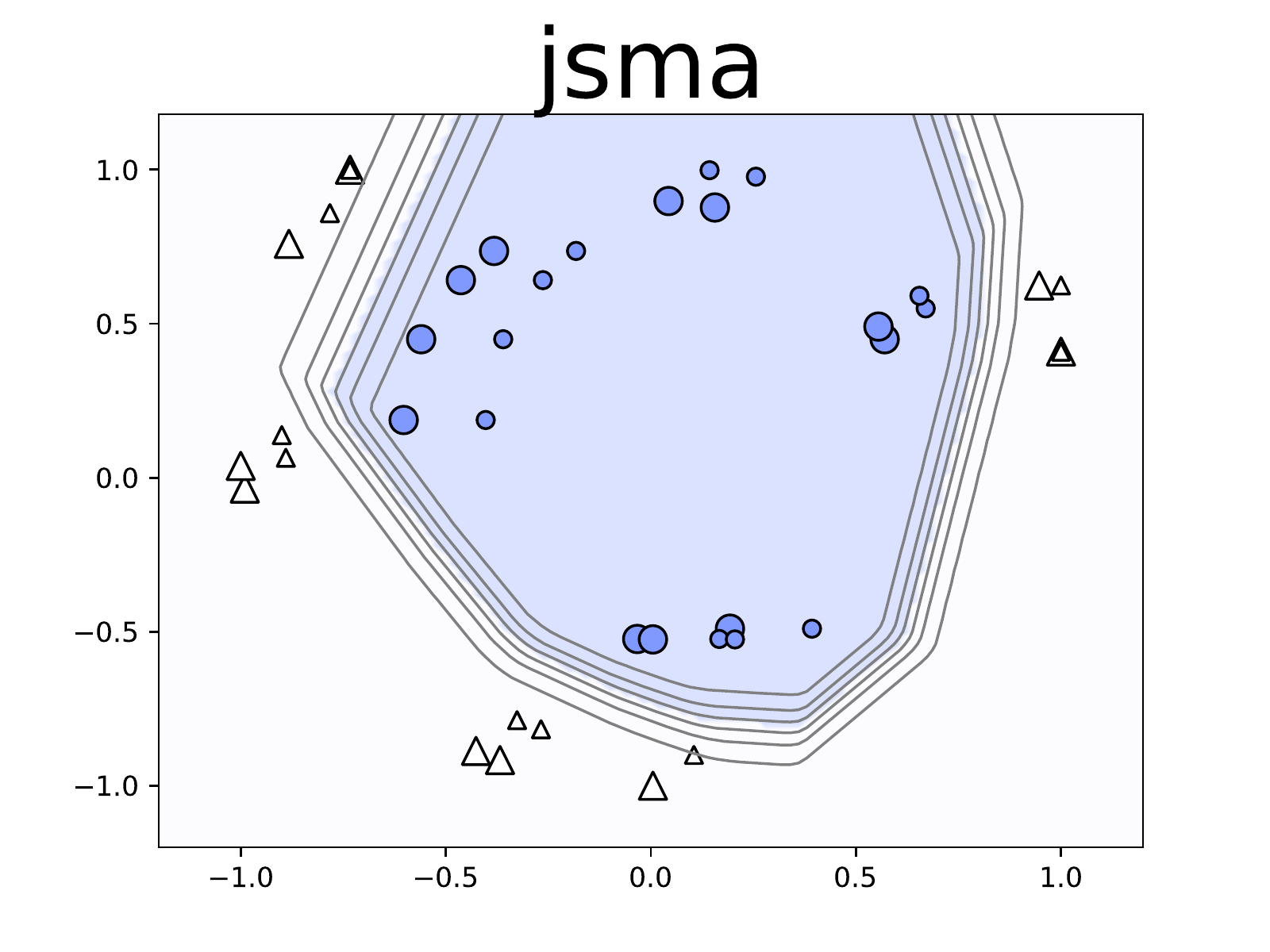}
     \caption{Concentric circles toy dataset.}
     \label{fig:circles}
   \end{subfigure}
   \begin{subfigure}{\textwidth}
     \centering
       \includegraphics[width=0.32\textwidth]{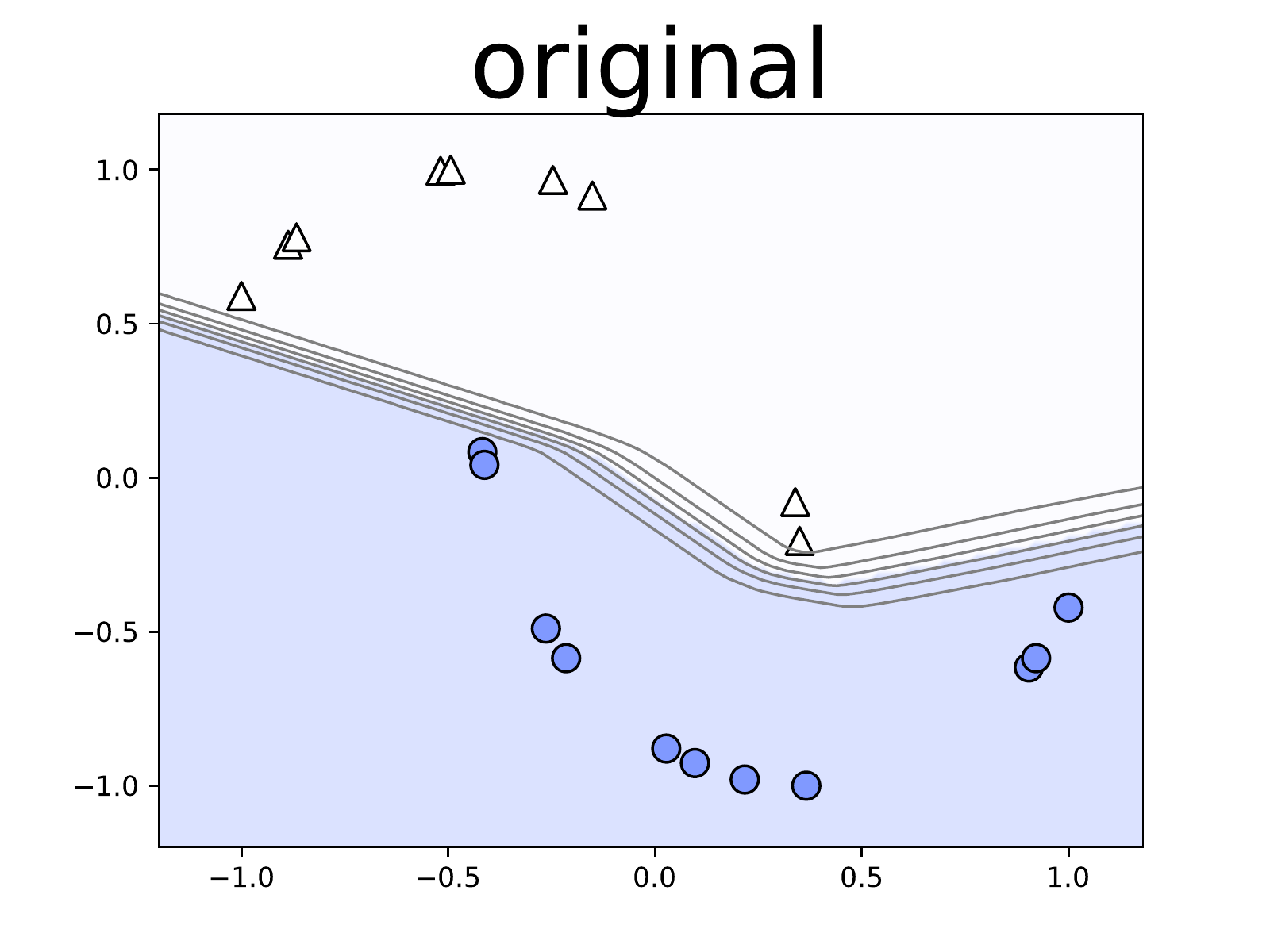}
       \includegraphics[width=0.32\textwidth]{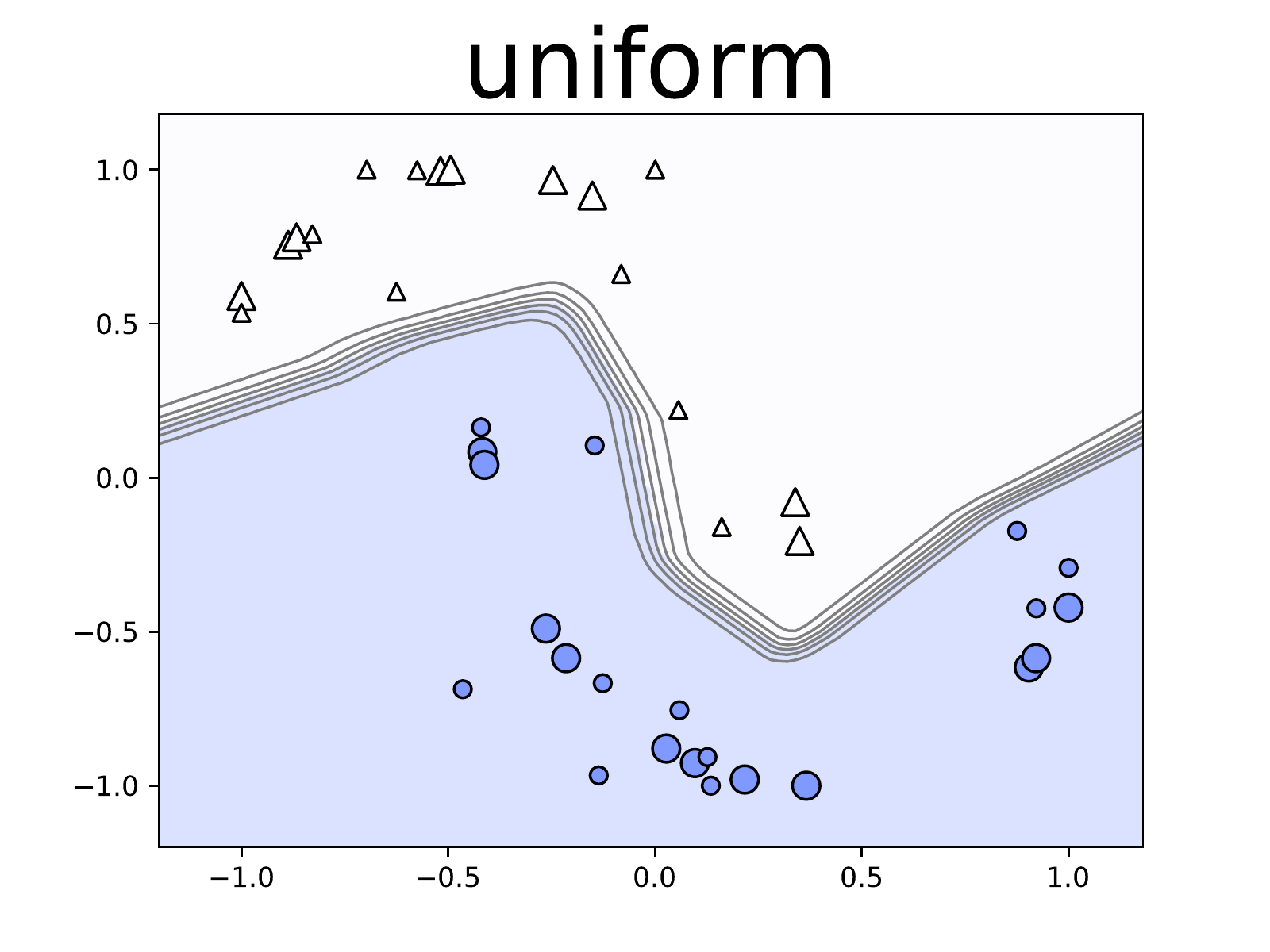}
       \includegraphics[width=0.32\textwidth]{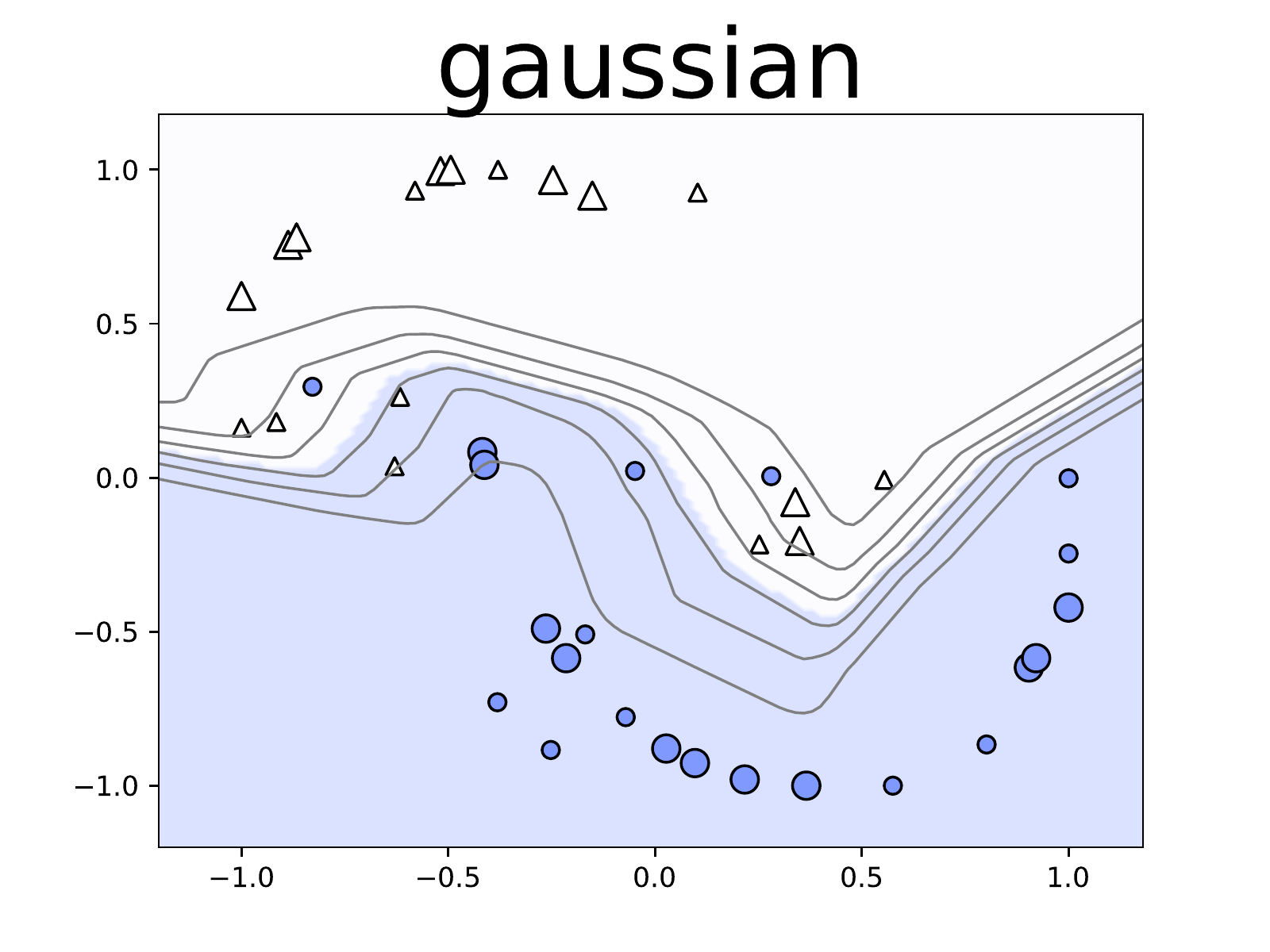}\\
       \includegraphics[width=0.32\textwidth]{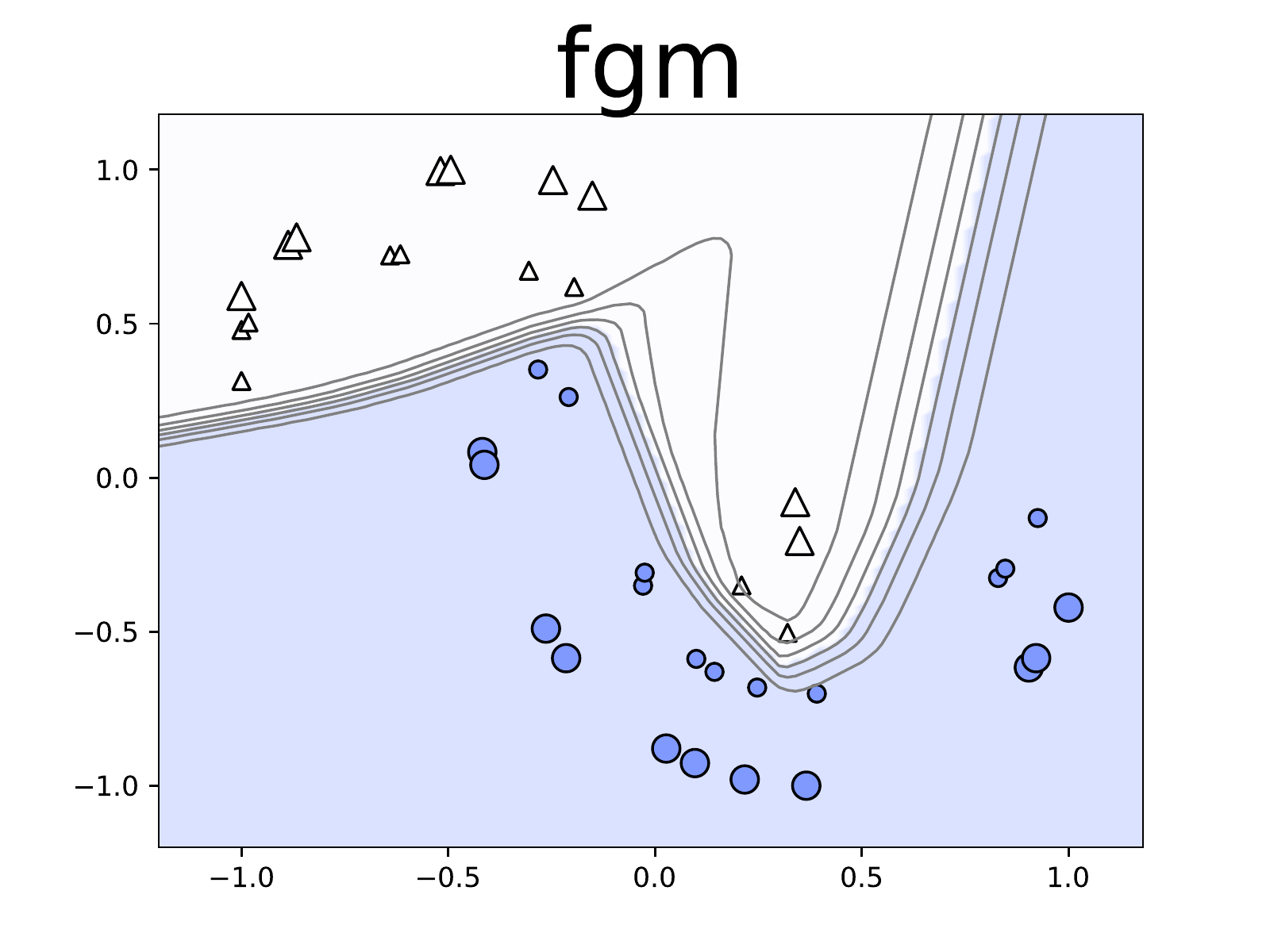}
       \includegraphics[width=0.32\textwidth]{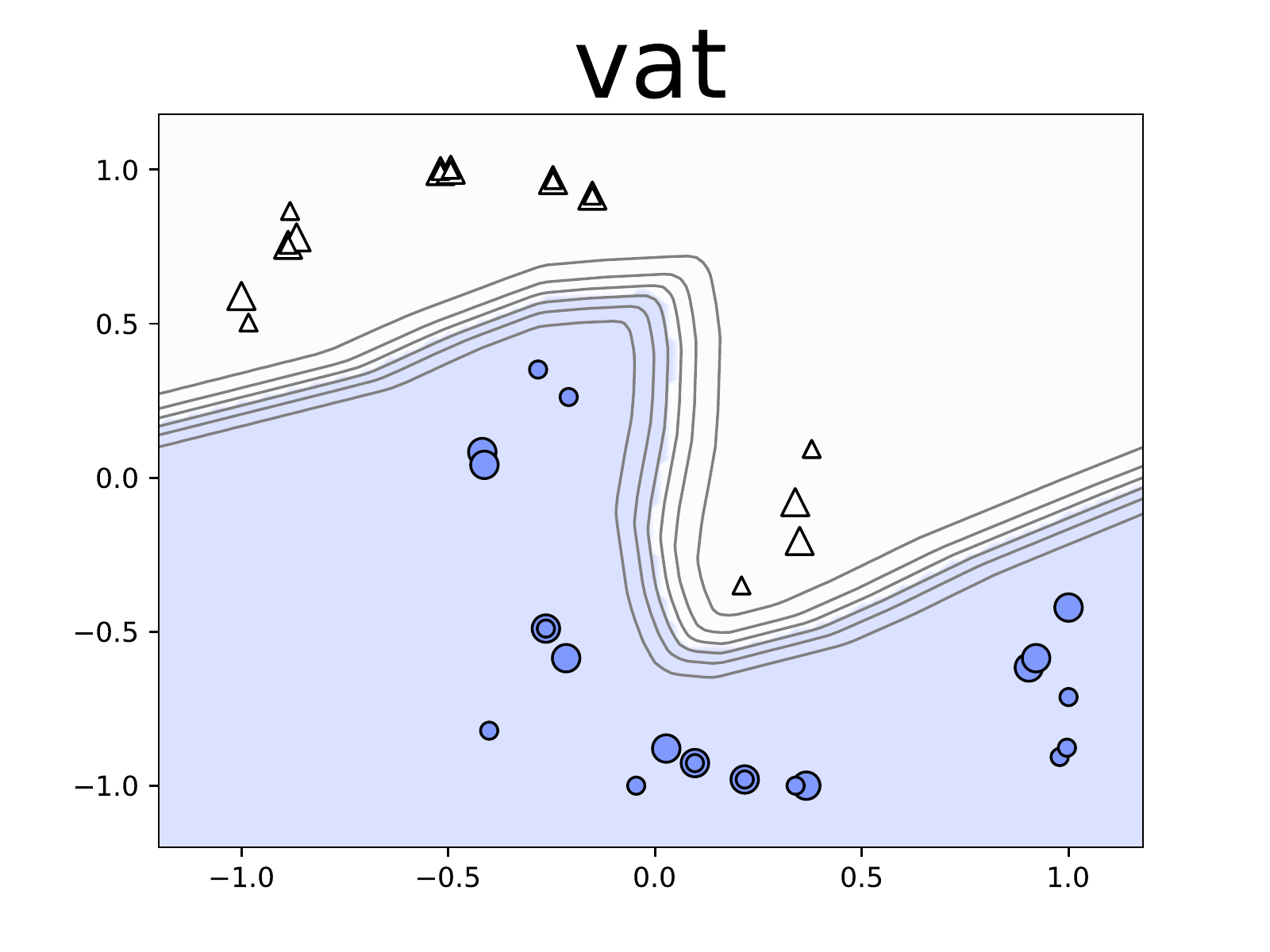}
       \includegraphics[width=0.32\textwidth]{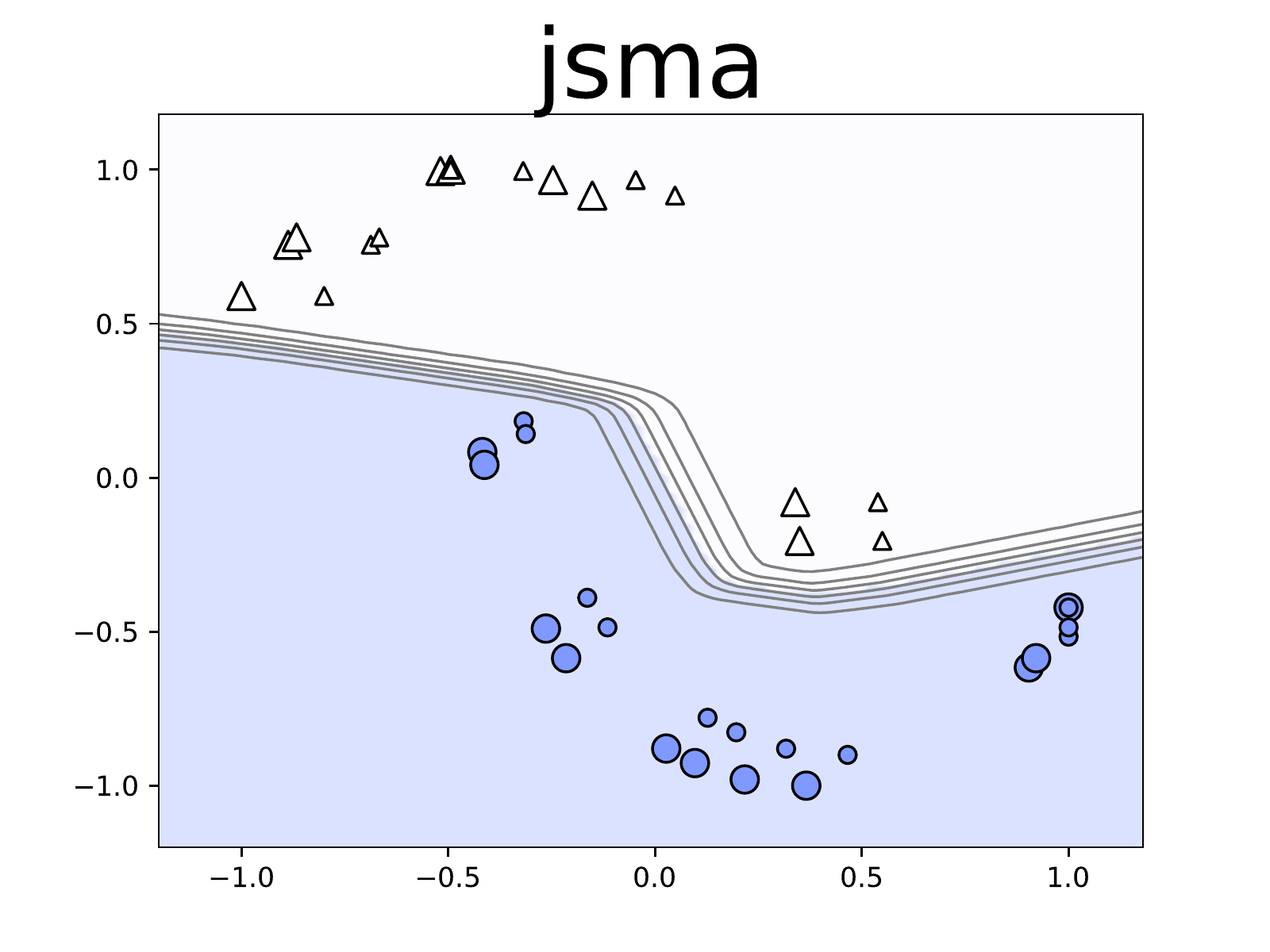}
      \caption{Half-moons toy dataset.}
      \label{fig:moons}
    \end{subfigure}
  \caption{Classification boundaries and confidence levels on toy datasets. We compare state-of-the-art data augmentation techniques for hardening learning models, in this case a soft-max neural network with two dense hidden layers and RELU activation function. The decision boundary can be identified through the colors and the confidence level contours are marked with black lines. The original points and the additional ones (smaller) are drawn with the label and color corresponding to their class.}
  \label{fig:toyset}
\end{figure}

To prove the benefits of the proposed Gaussian data augmentation (\textbf{GDA}) on the robustness of a model, we carry out a study on the classification boundaries and the confidence levels of a simple multi-layer network trained on two toy datasets augmented through different techniques.
In the first dataset (Figure~\ref{fig:circles}), the two classes are placed as concentric circles, one class inside the other.
The second dataset (Figure~\ref{fig:moons}) is the classic example of two half-moons, each representing one class.
% Plus, the model doesn't have to be trained twice and the computations are less expensive than crafting adversarial examples.
For each original point $x$ of a dataset, we add one of the following artificial points ${x}'$ (all of them clipped into the input domain):
\begin{enumerate}
  \item Adversarial example crafted with Fast Gradient Sign Method with $L_2$-norm and $\epsilon = 0.3$;
  \item Virtual adversarial example (VAT) with $\epsilon = 0.3$;
  \item Adversarial example crafted with Jacobian Saliency Method with feature adjustment $\theta=0.1$;
  \item Perturbed example drawn from the Uniform distribution centered in $x$ (${x}' \sim \mathcal{U}(x)$);
  \item Perturbed example drawn from a Gaussian distribution centered in $x$, with standard deviation $\sigma = 0.3$ (${x}' \sim \mathcal{N}(x,\sigma^2)$).
\end{enumerate}

GDA helps smoothing the model confidence without affecting the accuracy on the true examples (sometimes even improving it), as shown in Figure~\ref{fig:toyset}.
Notice how the change in the value of the loss function is smoother for GDA than for the other defense methods.
We also compare against augmentation with uniformly generated random noise, as was previously done in~\citep{miyato2017virtual}.
In this case as well, the GDA makes up for smoother variations.

These experiments confirm that, even though perturbing the true instances with random noise does not produce successful attacks~\citep{hazan2016perturbation}, it is highly effective when deployed as a defense.
Moreover, from a computational point of view, this new technique comes at practically no cost: the model does not require retraining, as opposed to adversarial training or defensive distillation, and drawing points from a Gaussian distribution is considerably cheaper than crafting adversarial examples.

% We join the two strategies proposed in this section into one defense method which we call [NONAME].
% Compared to adversarial training based on an attack, our defense has the major advantage of being computationally inexpensive, first because it only requires one training iteration, and second because Gaussian noise has much lower computational cost than crafting adversarial examples.
% The latter property allows us to reinforce a much higher number of directions around each input than adversarial training.
% The overall effect is obtaining a smoother, more stable model, which is able to sustain a wide range of adversarial attacks.
% As we will show in the experimental section, the original classification performance on clean input with our defense is maintained.

\section{Experiments}
\label{sec:experiments}
In this section, we discuss the experiments conducted for a closer look at the proposed defense methods, and contrast their performance against other defenses under multiple attacks.
Following a brief description of the experimental protocol, we detail the results obtained for an extensive class of setups.
The various settings and evaluation metrics are motivated to acquire better insights into the workings of adversarial misclassification and into the robustness of methods.
% We analyze the defense methods under multiple metrics in order to determine their robustness to adversarial perturbations.

\subsection{Setup}

\paragraph{Datasets}
Our experiments are performed on two standard machine learning datasets: MNIST~\citep{mnist} and CIFAR10~\citep{cifar}.
% Our experiments are performed on the standard MNIST dataset of handwritten digits.
MNIST contains 70,000 samples of black and white 28$\times$28 images, divided into 60,000 training samples and 10,000 test samples.
Each pixel value is scaled between 0 and 1, and the digits are the 10 possible classes.
CIFAR10 contains 60,000 images of size 32$\times$32 with three color channels, their values also scaled between 0 and 1.
The dataset is split into 50,000 training images and 10,000 test ones, all from 10 classes.
We consider two types of network architectures: simple convolutional neural nets (CNN)~\citep{lecun1998cnn} and convolutional neural nets with residuals (ResNet)~\citep{he2015resnet}.
The CNN is structured as follows: Conv2D(64, 8$\times$8) -- Conv2D(128, 6$\times$6) -- Conv2D(128, 5$\times$5) -- Softmax(10).
The ResNet has the following layers with an identity short-cut between layer $2$ and $6$: Conv2D(64, 8$\times$8) -- Conv2D(128, 6$\times$6) -- Conv2D(64, 1$\times$1) -- Conv2D(64, 1$\times$1) -- Conv2D(128, 1$\times$1) -- MaxPooling(3$\times$3) -- Softmax(10).
For both models, the activation function used is RELU, except when specified otherwise.

\paragraph{Methods} We use the following attacks to craft adversarial examples, in view of comparing the respective capacities of defense models:
\begin{enumerate}
  \item Fast Gradient Sign Method (FGSM) with distortion size $\epsilon$ ranging from 0.01 to 1, or with an iterative strategy determining the minimal perturbation necessary to change predictions;
  \item FGSM applied after a preliminary step of adding Gaussian noise (within a range of $\alpha=0.05$); the random noise applied in the first step is deducted from the budget of the FGSM attack; we call this heuristic Random + FGSM;
  \item Jacobian Saliency Map Attack (JSMA) with the default parameters from the authors' code: $\gamma=1$, $\theta=0.1$;
  \item DeepFool with a maximum of 100 iterations;
  \item The $L_2$ attack from~\citet{carlini2017} (called C\&W) with the default parameters from the authors' code and a confidence level of 2.3.
\end{enumerate}
We compare the following defense methods:

\begin{enumerate}
  \item Feature squeezing (FS), reducing the color depth to 1 bit for MNIST and 3 bits for CIFAR10, as the authors suggested in~\citet{xu2017feature_squeeze};
  \item Label smoothing (LS), with the weight of the true label set to $0.9$;
  \item Adversarial training (AT) with examples crafted using FGSM with $\epsilon=0.3$ for MNIST and $\epsilon=0.05$ for CIFAR10;
  \item Virtual adversarial training (VAT) with $\epsilon=2.1$ as indicated in~\citet{miyato2017virtual};
  \item Gaussian Data Augmentation generating ten noisy samples for each original one, with standard deviation $\sigma = 0.3$ for MNIST and $\sigma = 0.05$ for CIFAR10; when BRELU ($t=1$) is used as activation function, we call this method GDA + BRELU, otherwise GDA + RELU.
\end{enumerate}

Notice that we do not compare directly against defensive distillation, but consider label smoothing instead, for the reasons given in Section~\ref{sec:related}.
% For feature squeezing, we use a color depth of 1 bit, the same as the authors have done for MNIST.
% As virtual adversarial training is an effective defense, but not an effective attack, we only consider it as a defense.

\paragraph{Measuring defense efficiency}
Most experimental studies evaluate the performance of defense strategies on the basis of their effect on classification accuracy.
However, we believe it is also vital to measure the robustness of the obtained models to an attack~\citep{DBLP:journals/corr/Moosavi-Dezfooli15}, for instance by quantifying the average distortion introduced during adversarial generation.
These metrics provide better insights into the local behavior of defenses.
We compute three such metrics in our experiments: the empirical robustness (as measured by minimal perturbation), the distance to the training set and the loss sensitivity to pertubations.
For a model $F$, the robustness is defined as:

\begin{equation}
\rho_F = \mathbb{E}_{(x,y) \sim \mathcal{D}}\frac{\Delta x}{||x||_2},
\end{equation}
where $\Delta x$ amounts to the perturbation that is required for an instance $x$ in order for the model $F$ to change its prediction under a certain attack.
Intuitively, robust defenses require larger perturbations before the prediction changes.
We estimate robustness by its empirical value computed on the available sample.

Training set distance offers a complementary viewpoint to robustness.
In its attempt to quantify the dissimilarity between two sets, namely training data and adversarial images, this metric measures the average nearest-neighbour distance between samples from the two sets.
Consequently, a larger value of this metric for adversarial images obtained by minimal perturbation supports the robustness of a defense strategy.
Furthermore, this metric could also serve as a detection tool for adversarial attacks.

One of the main features of the proposed method is the smoothing effect on the learned model.
We propose to quantify this smoothness by estimating the Lipschitz continuity constant $\ell$ of the model, which measures the largest variation of a function under a small change in its input: the lower the value, the smoother the function.
In practice, we are unable to compute this theoretical metric.
We propose instead to estimate $\ell_F$ by local loss sensitivity analysis~\citep{krueger2017memory}, using the gradients of the loss function with respect to the $M$ input points in the test set:
\[
  \ell_F = \frac{1}{M} \sum_{i=1}^M \left|\left| \frac{\partial J(\theta, x_i, y_i)}{\partial x_i} \right| \right|_2.
\]

\subsection{Comparison Between Architectures}

% Remark:
% - adversarial training works well only when (1) applied on the same net architecture and (2) (supposition) using the same attack

\begin{figure}
  \centering
  \includegraphics[width=.55\textwidth]{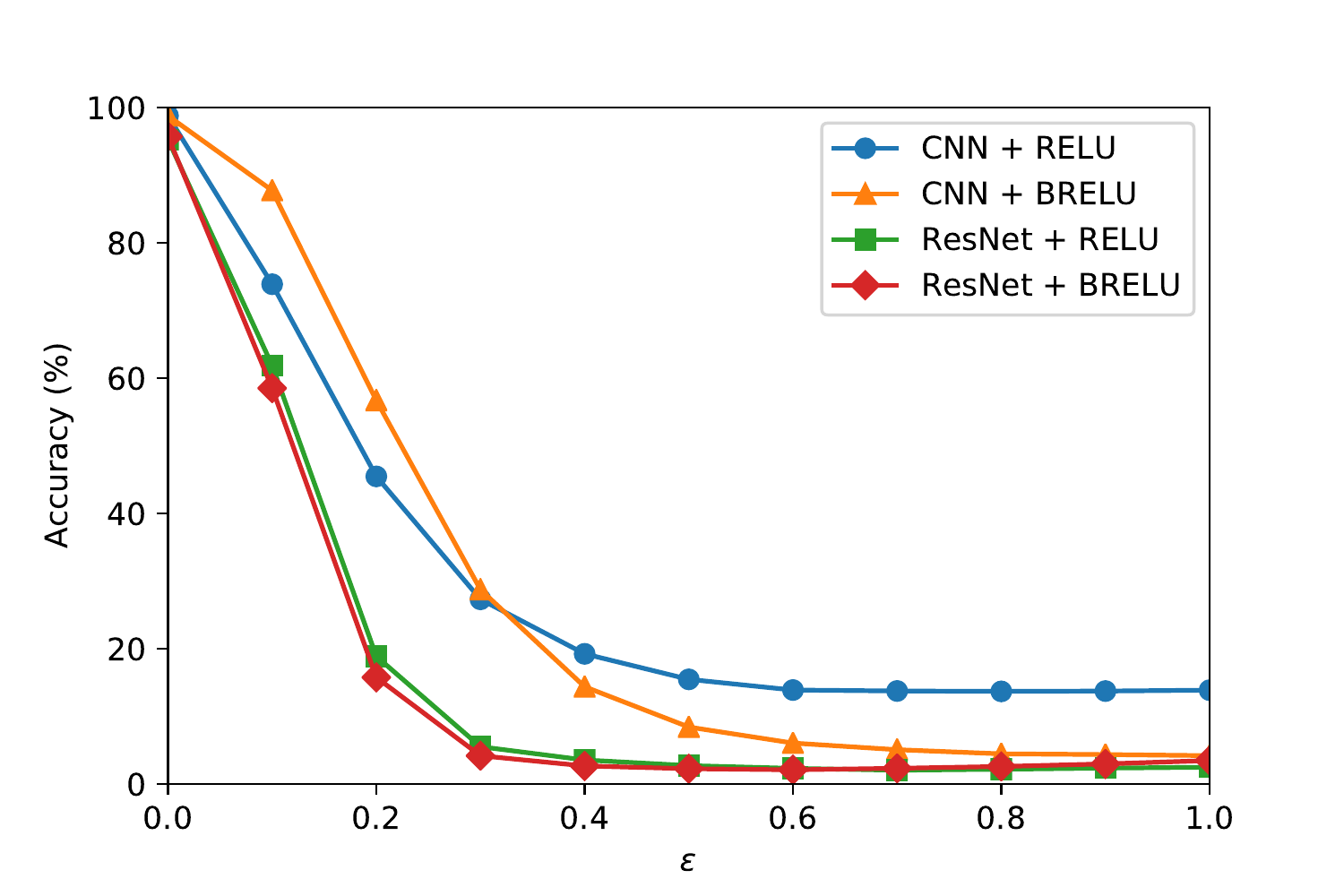}
  \caption{Accuracy on FGSM white-box attack with respect to $\epsilon$ for different architectures on MNIST.}
  \label{fig:archi_white_box}
\end{figure}

In this experiment, we aim to show the impact of the type of deep network architecture against adversarial attacks.
Namely, we compare CNN against ResNet, and RELU against BRELU activations respectively, under attacks crafted with FGSM.
As a matter of fact, we would like to reject the hypothesis of vanishing units, for which the misclassification of the adversarial examples would be caused by the deactivation of specific units and not by the activation of the wrong ones.
If that was the case, (i) getting residuals from the previous layers as in ResNet would result in higher robustness and (ii) using BRELU activation functions would decrease the performances of the model on adversarial examples.
However, as Figure~\ref{fig:archi_white_box} shows, ResNet does not sustain attacks better than CNN, suggesting that the accumulation of errors through the neural network is the main cause of misclassification.
Notice that CNN + BRELU performs best for a distortion of up to $\epsilon=0.3$, which has been suggested to be the highest value for which an FGSM attack might be undetectable.

% FGSM:
% - CNN versus ResNet and RELU versus BRELU
% - all $\epsilon$ for white-box $\rightarrow$ figure
% - fixed $\epsilon$ for black-box $\rightarrow$ table

% collcell or pgfplotstable+colortable for table gradient
\begin{table}
  \centering
  \caption{Accuracy (\%) on MNIST for FGSM attack transfer on different architectures ($\epsilon = 0.1$, best result in bold, second best in gray).}
  \label{tab:archi_black_box}
  \begin{small}
    \begin{tabular}{lcccc}
      \toprule
      \backslashbox{Crafted on}{Tested on} & CNN + RELU & CNN + BRELU & ResNet + RELU & ResNet + BRELU \\
      \midrule
      CNN + RELU      & 73.88           & 90.16          & \gray{89.49}   & \gray{88.26} \\
      CNN + BRELU     & 94.00           & 87.74          & \textbf{90.73} & \textbf{90.02} \\
      ResNet + RELU   & \textbf{94.46}  & \gray{93.65}   & 61.88          & 75.94 \\
      ResNet + BRELU  & \gray{94.21}    & \textbf{93.91} & 78.70          & 58.51 \\
      \bottomrule
    \end{tabular}
  \end{small}
\end{table}

We now analyze the same architectures for adversarial samples transferability: attacks crafted on each architecture with FGSM are applied to all the models (see Table~\ref{tab:archi_black_box}).
When the source and the target are the same, this makes for a white-box attack; otherwise, it is equivalent to a black-box setting.
As is expected, each architecture is most fooled by its own adversarial examples.
Additionally, ResNet is overall more affected by adversarial examples transferred from different architectures, while the CNN retains an accuracy higher than 90\% in all black-box attacks, even when the only change in the architecture is replacing RELU with BRELU.
This makes the case once more for noise accumulation over the layer rendering models vulnerable to adversarial samples.
Considering the results of the ResNet model, we perform the remaining experiments only on the simple CNN architecture.

\subsection{Impact of the Attack Distortion}
% 9 plots:
% - methods: FGSM
% - architectures: CNN (a -- white-box), ResNet (c -- black-box)
% - datasets: MNIST, CIFAR
% - curves per plot: vanilla, adversarial training, our method (GDA and GDA+BRELU)
% - 9th plot: attack on same architecture: adversarial training, GDA

\begin{figure}
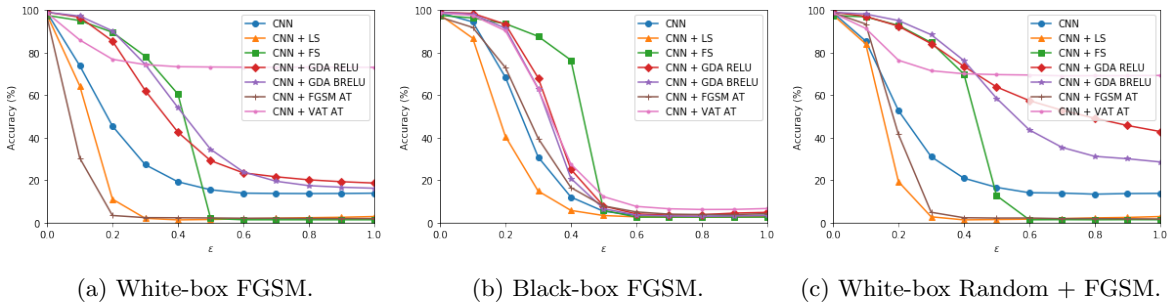

  \begin{subfigure}{.33\textwidth}
    \centering
    \includegraphics[width=\textwidth]{ex2/wb-fgsm}
     \caption{White-box FGSM.}
     \label{fig:ex2-wb-fgsm}
  \end{subfigure}
  \begin{subfigure}{.33\textwidth}
     \centering
       \includegraphics[width=\textwidth]{ex2/bb-fgsm}
      \caption{Black-box FGSM.}
      \label{fig:ex2-bb-fgsm}
  \end{subfigure}
  \begin{subfigure}{.33\textwidth}
    \centering
    \includegraphics[width=\textwidth]{ex2/wb-rndfgsm}
    \caption{White-box Random + FGSM.}
    \label{fig:wb-rndfgsm}
  \end{subfigure}
  \caption{Comparison of different defenses against white-box and black-box attacks on MNIST. For black-box attacks, the adversarial examples are crafted using the ResNet model, without any defense.}
  \label{fig:fgsm}
\end{figure}

\begin{figure}
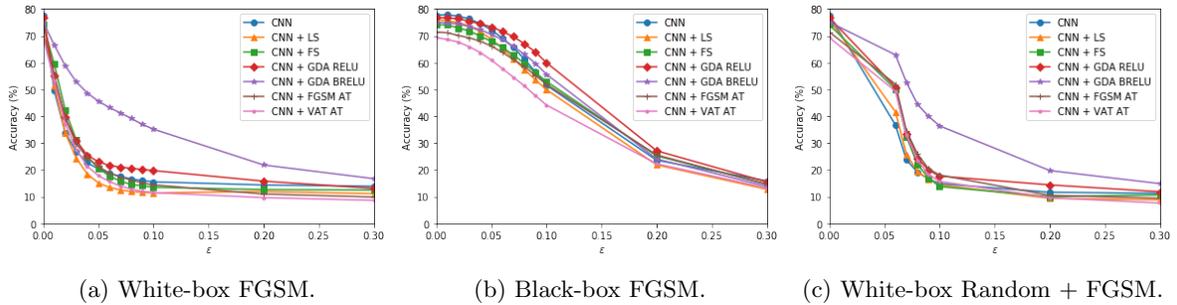

  \begin{subfigure}{.33\textwidth}
    \centering
    \includegraphics[width=\textwidth]{ex2/wb-fgsm-cifar}
     \caption{White-box FGSM.}
     \label{fig:cifar-wb-fgsm}
  \end{subfigure}
  \begin{subfigure}{.33\textwidth}
     \centering
       \includegraphics[width=\textwidth]{ex2/bb-fgsm-cifar}
      \caption{Black-box FGSM.}
      \label{fig:cifar-bb-fgsm}
  \end{subfigure}
  \begin{subfigure}{.33\textwidth}
    \centering
    \includegraphics[width=\textwidth]{ex2/wb-rndfgsm-cifar}
    \caption{White-box Random + FGSM.}
    \label{fig:cifar-wb-rndfgsm}
  \end{subfigure}
  \caption{Comparison of different defenses against white-box and black-box attacks on CIFAR10. For black-box attacks, the adversarial examples are crafted using the ResNet model, without any defense. Note that the adversarial examples crafted with Random + FGSM and $0 < \epsilon \leq 0.05$ correspond to Gaussian noise for $\alpha = 0.05$.}
  \label{fig:fgsm-cifar}
\end{figure}

% In , we study the impact of FGSM and Random + FGSM attacks on the accuracy of a CNN trained with different defenses.
We now study the impact of the attack distortion on the CNN model trained with different defenses.
To this end, we use FGSM and Random + FGSM as attacks and plot the results in Figures~\ref{fig:fgsm} and~\ref{fig:fgsm-cifar}.
Note that for CIFAR10, we only show the results for $0 \leq \epsilon \leq 0.3$, as for bigger values, the accuracies are almost constant.
The first observation is that label smoothing not only fails to strengthen the model in both setups, but it even worsens its robustness to adversarial attacks.
Also, adversarial training seems ineffective even against white-box attacks, contrary to the observations in~\cite{tramer2017ensemble}.
The result is not surprising, as the model is enforced on specific directions and probably loses its generalization capabilities.
We explain this only apparent contradiction by pointing out the difference in the definitions of a white-box attack: for~\cite{tramer2017ensemble}, white-box adversarial means examples crafted on the original model, without defenses; for us, they are crafted on the model trained through adversarial training.
For MNIST, the three defenses that consistently perform the best are feature squeezing, virtual adversarial training and Gaussian data augmentation.
However, if we focus on the results for $\epsilon < 0.3$ (which are the most significant, as big perturbations are easily detectable and result in rubbish examples), VAT is not as effective as the other two methods.
On CIFAR10, small perturbations easily compromise the accuracy of the models even when defended by the methods the most efficient on MNIST, notably feature squeezing and virtual adversarial training.
Moreover, using these last two defenses seems to degrade the performances on true examples and to fail in strengthening the model from black-box and white-box attacks alike.

In conclusion, our defenses outperform state-of-the-art strategies in terms of accuracy on these two datasets.

% \begin{figure}
%  \centering
%  \includegraphics[width=.5\textwidth]{ex2/wb-rndfgsm}
%  \caption{Accuracy on Random + FGSM white-box attack with respect to $\epsilon$ for different defenses.}
%  \label{fig:wb-rndfgsm}
% \end{figure}

\subsection{Defense Performance under Multiple Metrics}

As discussed previously, the accuracy is not a sufficient measure for evaluating the performance of a model, especially in the case of adversarial samples: an attack might make for an incorrect prediction with respect to the original label in most cases, but if the adversarial examples cannot be mistaken as legitimate inputs, the attack is arguably not effective.
For this reason, we propose to evaluate the robustness of defenses, as well as the shift of adversarial examples with respect to the clean data distribution.
The adversarial examples are crafted incrementally using FGSM with small perturbations, and the algorithm stops when the prediction changes.
The results in Table~\ref{tab:wb_fgsm0.1} show that GDA with RELU as activation function performs best in terms of accuracy, being on a par with VAT.
The robustness indicates the average amount of minimal perturbation necessary to achieve misclassification.
This measure proves that both versions of the proposed defense yield a more robust model, potentially making the adversarial examples visually detectable.
It is interesting to notice that feature squeezing and label smoothing actually decrease the robustness of the model.
The last column in the table measures the average (Euclidean) distance of adversarial samples to the closest training point: a higher values indicates a larger shift between the two distributions.
Here as well, GDA with both setups obtains much better results than the other defenses.
Coupled with the robustness results, this metric confirms that one cannot have resistance to adversarial samples without an overall robustness of models, which calls for generic model reinforcement methods like the one proposed in this paper.
Table~\ref{tab:gradients} presents the measure of the sensitivity of the loss function under small variations for all studied defense methods.
Gaussian augmentation provides the smoothest model with the RELU activation.
Notice that feature squeezing and label smoothing both induce higher gradients in the model, while the other defenses enforce a level of smoothness close to the one of the original unprotected model.

\begin{table*}
  \centering
  \caption{White-box attack on adversarial examples from FGSM with minimal perturbation (best result in bold, second best in gray).}
  \label{tab:wb_fgsm0.1}
  \begin{tabular}{lccc}
    \toprule
    Defense & Accuracy & Robustness & Training set distance \\
    \midrule
     CNN                      & 52.07          & 0.202          &  3.90 \\
     Feature squeezing        & 35.61          & 0.143          &  3.01 \\
     Label smoothing          & 37.06          & 0.152          &  3.18 \\
     FGSM adversarial training & 56.44         & 0.226          &  4.33 \\
     VAT                      & \gray{73.32}   & 0.308          &  6.64 \\
     GDA + RELU               & \textbf{73.96} & \gray{0.440}   & \textbf{10.36} \\
     GDA + BRELU              & 69.56          & \textbf{0.471} & \gray{9.31} \\
    \bottomrule
  \end{tabular}
\end{table*}

% \begin{table}
%   \centering
%   \caption{White-box attack on adversarial examples from DeepFool with minimal perturbation.}
%   \label{tab:wb_deepfool0.1}
%   \begin{tabular}{lccc}
%     \toprule
%     Defense & Accuracy & Robustness & Training set distance \\
%     \midrule
% CNN                      & 79.16 &   0.155 &   2.31 \\
% Feature squeezing        & 97.23 &   0.179 &   2.35 \\
% Label smoothing          & 73.13 &   0.147 &   2.26 \\
% FGSM Adversrial Training & 98.26 &   0.026 &   2.88 \\
% VAT                      & 98.41 &   0.021 &   2.90 \\
% GDA + RELU               & 97.07 &   0.105 &   2.76 \\
% GDA + BRELU              &  9.97 & nan     & nan       \\
%    \bottomrule
%  \end{tabular}
% \end{table}

\begin{table}
  \centering
  \caption{Local loss sensitivity analysis for defenses on MNIST (best result in bold, second best in gray).}
  \label{tab:gradients}
  \begin{tabular}{lc}
    \toprule
    Model & Local sensitivity \\
    \midrule
    CNN                         & \gray{0.0609} \\
    Feature squeeze             & 0.1215 \\
    Label smoothing             & 0.2289 \\
    FGSM adversarial training   & 0.0748 \\
    VAT                         & 0.0741 \\
    GDA + RELU                  & \textbf{0.0244} \\
    GDA + BRELU                 & 0.0753 \\
    \bottomrule
  \end{tabular}
\end{table}

% \begin{table}
%   \centering
%   \caption{Local sensitivity analysis for all defenses on CIFAR10.}
%   \label{tab:gradients}
%   \begin{tabular}{lc}
%     \toprule
%     Model & Local sensitivity \\
%     \midrule
%     CNN                         & 1.5716 \\
%     Feature squeeze             & 0.9608 \\
%     Label smoothing             & 1.1460 \\
%     FGSM adversarial training   & 0.8296 \\
%     VAT                         & 0.6958 \\
%     GDA + RELU                  & 1.2134 \\
%     GDA + BRELU                 & 1.7108 \\
%     \bottomrule
%   \end{tabular}
% \end{table}

\subsection{Transferability of Adversarial Samples}

We now compare the performance of the proposed method against other defenses in a black-box setting, to account for the adversarial examples transferability phenomenon.
To this end, all adversarial examples are crafted on the ResNet model described previously, before being applied to the CNN model trained with each of the defense methods.
In the case of the FGSM attack, $\epsilon$ is set to 0.1.
Tables~\ref{tab:bb_fgsm0.1} and~\ref{tab:bb_cifar} present the classification accuracy obtained by the models.
The first line is the baseline, that is the CNN with no defense.
For our method, we consider GDA both with and without the use of BRELU.
On MNIST, our methods outperform the other defenses in most cases.
Namely, notice that they obtain the best performance under FGSM, VAT, DeepFool, JSMA and C\&W attacks; this difference is significant for FGSM, DeepFool and C\&W.
Random + FGSM has been shown to be a stronger attack than one-step attacks; this is confirmed by the poor performance of all defenses, except for feature squeezing.
An interesting fact is that using label smoothing degrades the performance of the model under the Random + FGSM attack when compared to the CNN with no defense.
On CIFAR10, our methods consistently make the model more robust than the other defenses.
% These results suggest that DeepFool, JSMA and C\&W are weaker attacks than FGSM based ones, in the studied black-box setting.
% all strategies could seem ineffective in defending the model.
% However the accuracy of our CNN classifier on the true dataset ($77.72 \%$) is close to those values.

\begin{table*}
  \centering
  \caption{Accuracy (\%) for black-box attacks on MNIST (best result in bold, second best in gray).}
  \label{tab:bb_fgsm0.1}
  \begin{tabular}{lccccc}
    \toprule
    \backslashbox{Defense}{Attack} & FGSM & Rand + FGSM & DeepFool & JSMA & C\&W \\
    \midrule
 CNN                       & 94.46          & 40.70          & 92.95          & 97.95          & 93.10 \\
 Feature squeezing         & 96.31          & \textbf{91.09} & 96.68          & 97.48          & 96.75 \\
 Label smoothing           & 86.79          & 20.28          & 84.58          & 95.86          & 84.81 \\
 FGSM adversarial training & 91.86          & 49.77          & 85.91          & 98.62          & 97.71 \\
 VAT                       & 97.53          & 74.35          & 96.03          & 98.26          & 96.11 \\
 GDA + RELU                & \textbf{98.47} & \gray{80.25}   & \gray{97.84}   & \textbf{98.96} & \gray{97.87} \\
 GDA + BRELU               & \gray{98.08}   & 75.50          & \textbf{98.00} & \gray{98.88}   & \textbf{98.03} \\
    \bottomrule
  \end{tabular}
\end{table*}

\begin{table*}
  \centering
  \caption{Accuracy (\%) for black-box attacks on CIFAR10 (best result in bold, second best in gray).}
  \label{tab:bb_cifar}
  \begin{tabular}{lccccc}
    \toprule
    \backslashbox{Defense}{Attack} & FGSM & Rand + FGSM & DeepFool & JSMA & C\&W \\
    \midrule
    CNN                       & 51.55          & 57.22          & \textbf{77.76} & \textbf{76.85} & \textbf{77.79} \\
    Feature squeezing         & 52.98          & 61.83          & 73.82          & 73.28          & 73.75 \\
    Label smoothing           & 49.90          & 56.68          & 75.80          & 74.78          & 75.66 \\
    FGSM adversarial training & 51.94          & 63.25          & 71.17          & 70.48          & 71.02 \\
    VAT                       & 44.16          & 56.47          & 68.94          & 68.21          & 68.47 \\
    GDA + RELU                & \textbf{59.96} & \textbf{71.93} & \gray{76.80}   & \gray{76.45}   & \gray{76.61} \\
    GDA + BRELU               & \gray{55.65}   & \gray{69.43}   & 74.97          & 74.73          & 74.78 \\
    \bottomrule
  \end{tabular}
\end{table*}

% \subsection{S}
% In this final experiment, we study the scalability of GDA on a dataset of high dimensionality, ImageNet~\citep{0f810e97f6074261b823b192b8084243}.
% We show that
% Instead of training a model from scratch, we fine-tune VGG16~\citep{simonyan2014very} pre-trained on ImageNet.

\section{Conclusion}
\label{sec:conclusion}

Despite their widespread adoption, deep learning models are victims of uninterpretable and counterintuitive behavior.
While numerous hypotheses compete to provide an explanation for adversarial samples, their root cause still remains largely unknown.
The quest of understanding this phenomenon has turned into an arms race of attack and defense strategies.
Along with the drive for efficient and better attacks, there is a parallel hunt for effective defenses which can guard against them.
Given this large ammo of strategies, practitioners are faced with a dire need for attack-agnostic defense schemes which can be easily employed at their end.

This work proposed two such strategies which, used separately or combined, improve the robustness of a deep model.
We built on two intuitive hypotheses of error accumulation and smoothness assumption, and proposed to impose two constraints: first, on the architecture in the form of the bounded RELU activations, and second, in the form of training with Gaussian augmented data.
We demonstrated the utility of this combined approach against the state-of-art attacks.
Compared to adversarial training based on an attack, our defense has the major advantage of being computationally inexpensive; first, it only requires training one model, and second, Gaussian noise has much lower computational cost than crafting adversarial examples.
The latter property allows us to explore a larger set of directions around each input than adversarial training.
The overall effect is obtaining a smoother, more stable model, which is able to sustain a wide range of adversarial attacks.
As we have shown in the experimental section, we achieve this without compromising on the original classification performance on clean input.

\bibliographystyle{plainnat}
\begin{small}
\bibliography{references}
\end{small}
\end{document}